\documentclass[conference]{IEEEtran}
\pdfoutput=1
\usepackage{url}
\usepackage{subfig}
\usepackage{cite}
\usepackage{amsmath,amssymb,amsfonts}
\usepackage{algorithmic}
\usepackage{graphicx}
\usepackage{textcomp}
\usepackage{adjustbox}
\usepackage{graphicx}
\usepackage{xcolor}
\usepackage{mathtools,lipsum, nccmath,cuted}
\usepackage{siunitx}
\usepackage{textcomp}

\usepackage{comment}
\usepackage{makecell}
\usepackage{algorithm}
\usepackage{algorithmic}
\usepackage{gensymb}
\usepackage[normalem]{ulem}
\usepackage{colortbl}
\usepackage{tabularx}
\usepackage{array}
\newcolumntype{L}{>{\centering\arraybackslash}m{3cm}}
\usepackage{enumitem}
\definecolor{Gray}{gray}{0.925}
\usepackage[textsize=tiny]{todonotes}
\usepackage[utf8]{inputenc}
\usepackage[acronym]{glossaries}
\usepackage{longtable}
\usepackage{acro}
\usepackage{balance}
\usepackage{ragged2e}
\usepackage[capitalise]{cleveref}

\acsetup{first-style=short,list-style=tabular}
\DeclareAcronym{alc}{
	short = ALC,
	long = Assurance-based Learning-enabled CPS,
	class = abbrev
}
\DeclareAcronym{cps}{
	short = CPS,
	long = Cyber Physical System,
	class = abbrev
}
\DeclareAcronym{gsn}{
	short = GSN,
	long = Goal Structuring Notation,
	class = abbrev
}
\DeclareAcronym{acg}{
	short = ACG,
	long = Assurance Case Generation,
	class = abbrev
}
\DeclareAcronym{aebs}{
	short = AEBS,
	long = Automatic Emergency Braking System,
	class = abbrev
}
\DeclareAcronym{sc}{
	short = SC,
	long = Safety Case,
	class = abbrev
}

\DeclareAcronym{cc}{
	short = CC,
	long = Certification Criteria,
	class = abbrev
}

\DeclareAcronym{ac}{
	short = AC,
	long = Assurance Case,
	class = abbrev
}

\graphicspath{ {figures/} }
\def\BibTeX{{\rm B\kern-.05em{\sc i\kern-.025em b}\kern-.08em
    T\kern-.1667em\lower.7ex\hbox{E}\kern-.125emX}}
    
\usepackage{pgfplots}
\usepgfplotslibrary{colorbrewer}
\pgfplotsset{compat = 1.14, cycle list/Set1-8} 
\usetikzlibrary{pgfplots.statistics, pgfplots.colorbrewer,pgfplots.dateplot,
        shadows,
        matrix} 

\usepackage{pgfplotstable}
\usepackage{filecontents}
\usepackage{graphicx}
\pgfplotsset{compat=1.14}
\definecolor{blueLine}{RGB}{57,106,177}
\definecolor{blueFill}{RGB}{114,147,203}
\definecolor{redLine}{RGB}{204,37,41}
\definecolor{greenline}{RGB}{0,250,0}
\definecolor{blackLine}{RGB}{0,0,0}
\definecolor{goldLine}{RGB}{160,82,45}
\usepackage[font=footnotesize]{caption}

\usepackage{soul}
\usepackage{xcolor}

\usepackage[font=footnotesize]{caption}

\begin{document}

\title{
{A Methodology for Automating Assurance Case Generation}}

\author{\IEEEauthorblockN{{Shreyas Ramakrishna\IEEEauthorrefmark{1}, Charles Hartsell\IEEEauthorrefmark{1}, Abhishek Dubey\IEEEauthorrefmark{1}, Partha Pal\IEEEauthorrefmark{2},
Gabor Karsai\IEEEauthorrefmark{1}}}
\IEEEauthorblockA{\IEEEauthorrefmark{1}\textit{Institute for Software Integrated Systems}\\
\textit{Vanderbilt University}\\
\textit{Nashville, TN, USA}
}
\IEEEauthorblockA{\IEEEauthorrefmark{2}\textit{Raytheon BBN Technologies}\\
\textit{Cambridge, MA 02138}
}
}



\pagestyle{plain}

\newpage

\maketitle

\begin{abstract} 
Safety Case has become an integral component for safety-certification in various Cyber Physical System domains including automotive, aviation, medical devices, and military. The certification processes for these systems are stringent and require robust safety assurance arguments and substantial evidence backing. Despite the strict requirements, current practices still rely on manual methods that are brittle, do not have a systematic approach or thorough consideration of sound arguments. In addition, stringent certification requirements and ever-increasing system complexity make ad-hoc, manual assurance case generation (ACG) inefficient, time consuming, and expensive. To improve the current state of practice, we introduce a structured ACG tool which uses system design artifacts, accumulated evidence, and developer expertise to construct a safety case and evaluate it in an automated manner. We also illustrate the applicability of the ACG tool on a remote-control car testbed case study.

\end{abstract}


{\footnotesize
\printacronyms[include-classes=abbrev, name=\textbf{ABBREVIATIONS}]}

\section{INTRODUCTION}
\label{sec:introduction}
Design of Assurance Case (\ac{ac}) for safety critical \acp{cps} has become an important industrial requirement. An assurance case \cite{bishop2000methodology,chowdhury2017principles} is a structured argument which composes different pieces of evidence to show that system-level goals have been satisfied. The goals (or claims) refer to the property of the system being monitored like safety, reliability, security, etc., and evidences are facts about a component of the system, that are accumulated by prior research or experiments. An argument links relevant evidences to the claims, and can either be deterministic, probabilistic or qualitative. Safety Case (\ac{sc}) is a specialized assurance case widely used for assuring the system level safety in CPS domains like aviation \cite{denney2012automating}, military \cite{kritzinger2006aircraft} and automotive \cite{palin2010assurance}. 





Despite its importance, construction of a safety case is often done manually by developers without a systematic approach or thorough consideration of safety arguments. Conventional safety case reports are long textual arguments which typically try to communicate and argue about the safety of a system. However, unclear and poorly structured textual language (mostly English) has always been a problem (explained in \cite{kelly2004systematic}) in communicating safety arguments among the different designers or operators involved. To overcome the irregularities with conventional textual safety reports and to make the documentation of safety case easy to read, graphical structures such as Claims-Argument-Evidence (CAE) \cite{bloomfield1998ascad} or Goal Structuring Notation (GSN) \cite{kelly2004systematic} was introduced. 

The GSN is widely used among the two. It is a graphical way of constructing a safety case using individual elements of claims, sub-claims, assumptions, arguments and evidences. The idea behind the goal structure is to show how the top-level claim or goal can be decomposed into multiple sub-goals connected with some argument structure. This process is repeated to further decompose sub-goals until it reaches the leaf-goals which can be directly supported by low-level component evidences. This graphical structuring of arguments has certainly simplified and improved the comprehension of assurance arguments across all stakeholders, thus encouraging its use in various safety-critical industries (listed in \cite{kelly2004systematic}). However, as explained in \cite{nair2015evidential}, GSN has only provided a simplified means of expressing arguments but has not improved the quality of the argument structure itself. Also, the design of the GSN is still performed manually by human experts which makes it time consuming and error prone.

There are several tools  \cite{steele2011access,adelard2011assurance,matsuno2011d,barry2011certware} that support development of CAE or GSN for safety cases. These tools provide excellent visual aid and editors for rapid prototyping, and management of large safety cases. However, they do not provide a mechanism to automate the construction of the CAE or GSN. These tools still require a human expert to manually generate the safety arguments that connect the goals to the low-level evidences. Also, as discussed in \cite{denney2018tool}, these tools lack a formal basis, which has limited their automation capability. Additionally, most of these tools do not provide a mechanism to quantify the confidence of the developed safety case, which is an important functionality. These limitations motivate the need for an automated tool that can develop and evaluate safety case with little human involvement.   


\begin{figure}[t]
\centering
\setlength{\belowcaptionskip}{-10pt}
 \includegraphics[width=0.9\columnwidth]{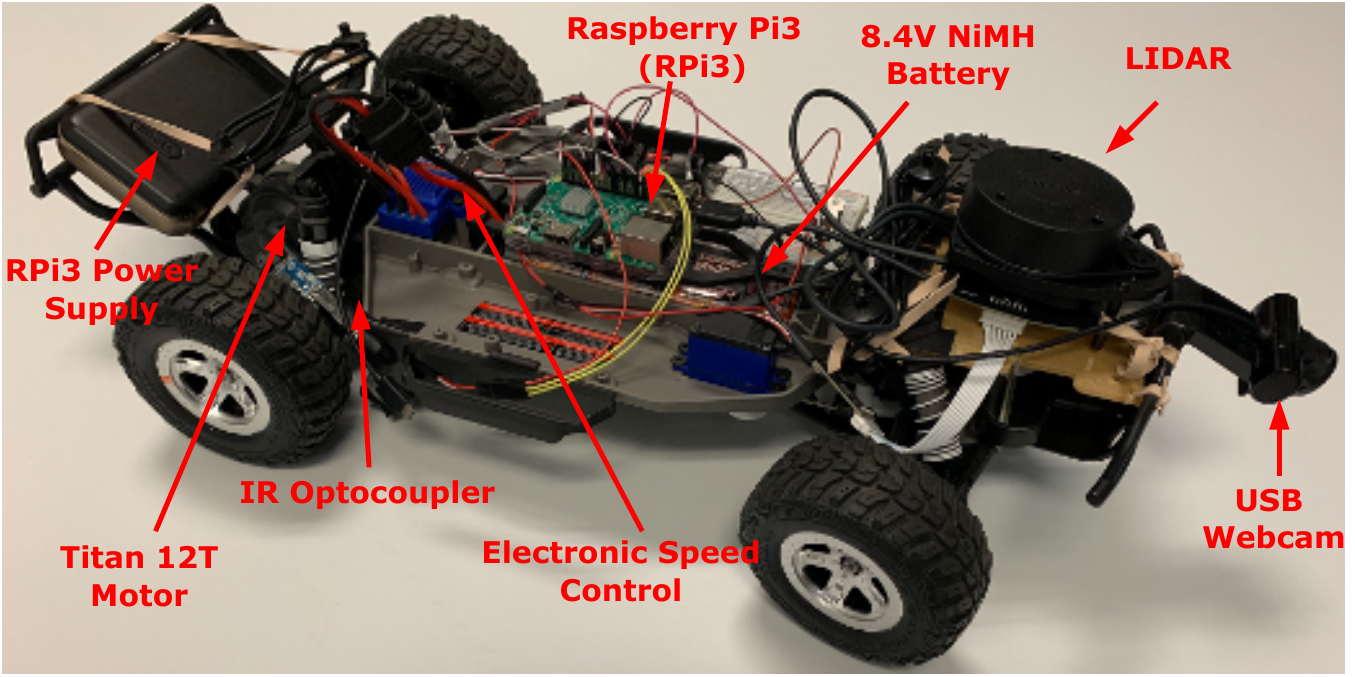}
 \caption{DeepNNCar is a resource-constrained autonomous RC car that uses Camera, LIDAR, IR-Optocoupler, and Raspberry Pi.}
 \label{fig:DeepNNCar}
 \end{figure}
 
 \begin{figure*}[t!]
    \centering
    \includegraphics[width=0.8\textwidth]{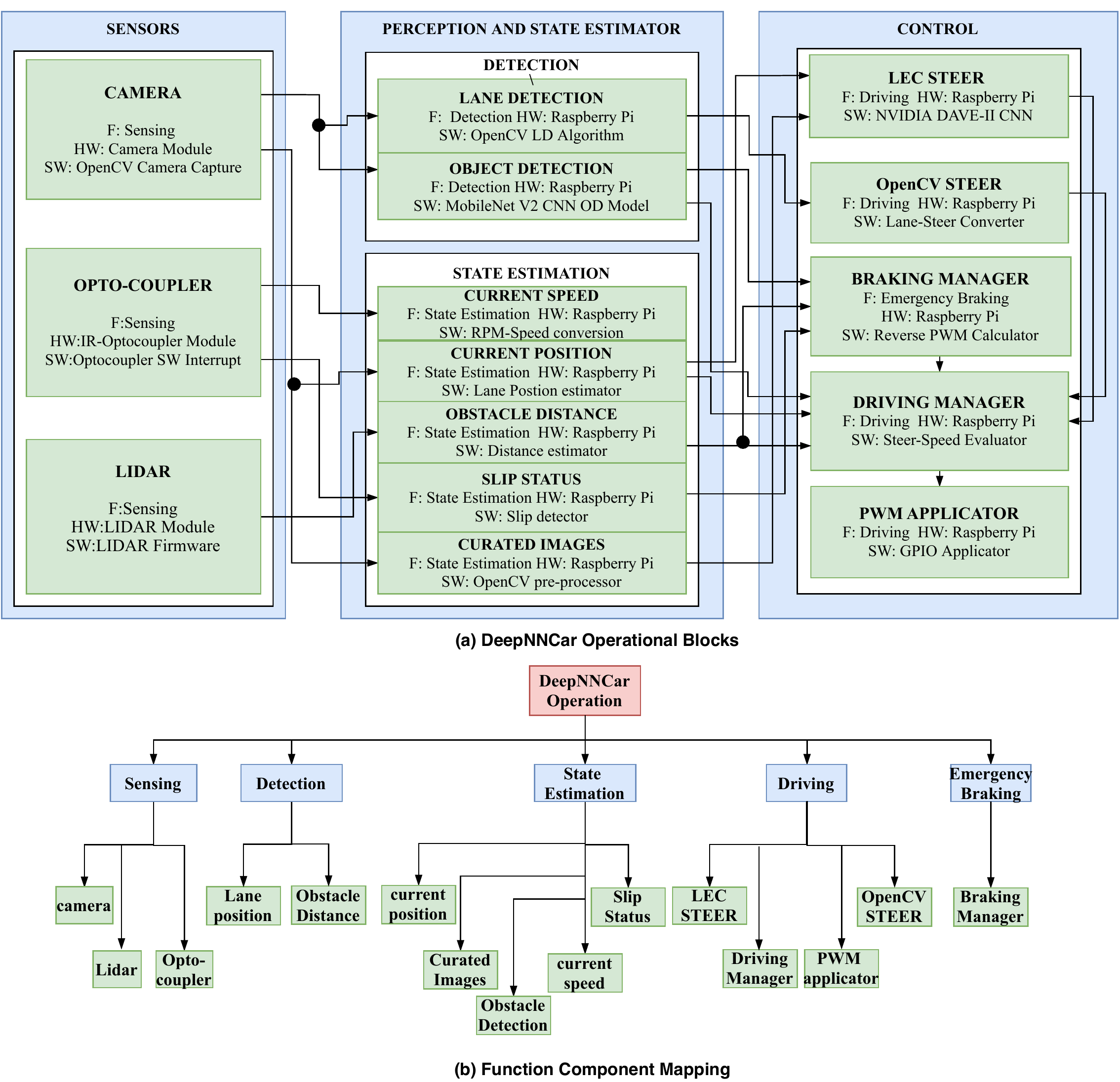}
    \caption{(a) The operation of the DeepNNCar is split into sensing-state estimation-control tasks. F -- represents the functionality of the components in the system, HW -- represents the hardware component which is responsible for the function, and SW -- represents the software component which is responsible for the function. The arrows between the tasks represents the interdependence of the different components and their functionality. (b) Function-component mapping of DeepNNCar operations. Shows the different hardware and software components that contribute to a function. }
     \label{fig:car_operation}
    \vspace{-0.6em}
\end{figure*}  



\textbf{Our Contributions}: In this work, we introduce an ACG tool that uses design artifacts of the target system along with the evidences aggregated by human experts to generate an safety case GSN starting from a given certification criteria (CC). Our goal with the ACG tool is to provide a structured method for the construction and evaluation of an assurance argument in support of a given certification criteria while automating the process to reduce human involvement. To realize these goals, the tool uses a classic divide and conquer strategy leveraging iterative search to decompose complex GSN goals and to find evidences that match them. We follow a rapid prototyping approach to realize the components of the ACG tool. Specifically, our contributions are as follows:
\begin{itemize}[noitemsep,leftmargin=*]
    \item We elaborate the three-step methodology of the ACG tool that automates the safety case generation process.
    \item We discuss three core steps of Link-Seal-Expand for iterative goal decomposition and automated GSN construction.
    \item We discuss a method using confidence score to evaluate the credibility of the generated safety case.
\end{itemize}

We evaluate the ACG tool for an illustrative example of an \ac{aebs} for a RC car testbed called DeepNNCar \cite{ramakrishna2019augmenting}. We hypothesize this automated safety assurance construction method would reduce the cost of assuring \ac{cps}, provide a robust process that minimizes human involvement, and accelerate the entire process of safety case generation. 

\textbf{Outline:} In \cref{sec:RW}, we discuss a few background concepts required to understand this work. \cref{sec:EBS} sets up an example of AEBS, that is used throughout the paper to explain the ACG concepts. \cref{sec:ACG} describes the core components of the ACG tool. In \cref{sec:example}, we illustrate the utility of ACG tool in the context of the AEBS example. \cref{sec:quality} discusses a safety case evaluation scheme using confidence score as a metric. \cref{sec:related_research} discusses the related work. In \cref{sec:discuss}, we make a brief discussion about the capabilities of the ACG tool, and thereafter we list a few possible future enhancements. Finally, in \cref{sec:conclusion} we present our conclusion.

\section{BACKGROUND}
\label{sec:RW}
This section provides an overview of safety case, GSN and the \ac{alc} Toolchain that is going to be extensively referred throughout this paper.

\subsection{Safety Case (SC)}
Safety Case has become an important requirement to prove the functional safety of safety-critical CPS, and has become a wide standard in the automotive, aerospace, and military industries. Safety case is a structured argument made using evidences to support the different claims made about the properties of the system. For every component or subsystem, a claim can be made about its safety, reliability, availability, security, etc. Then either deterministic, probabilistic, or qualitative arguments can be used along with evidences to prove if the claim about the system holds. The elements of a safety case as explained in \cite{bishop2000methodology} are:

\begin{itemize}[noitemsep,leftmargin=*]
    \item {Claim}: The property of a system that requires assurance (e.g. safety, reliability, availability, security, etc.)
    \item {Evidence}: Facts, sub-goals, assumptions, functions, and sub-arguments which provide a conclusive support to prove the claims made about the system.
    \item {Arguments}: A linking structure between the claim and the supporting evidence. Used to show how the claims are backed by the evidence available for the system.
    \item {Inference/Operators}: Provides rules or mechanisms to transform the arguments. 
\end{itemize}

\begin{figure*}[t!]
    \centering
    \includegraphics[width=0.85\textwidth]{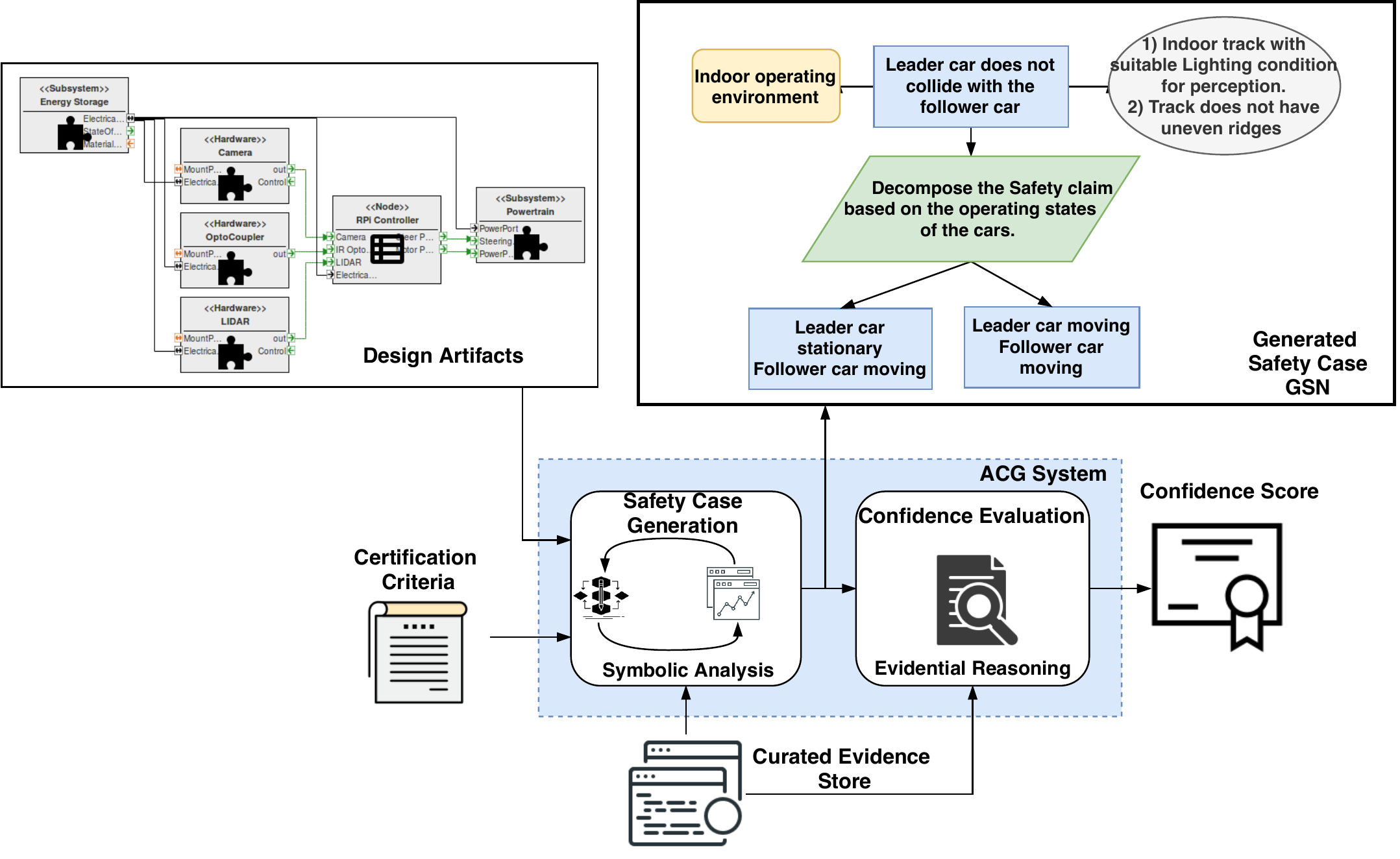}
    \caption{The high level approach of Assurance Case Generation tool, that uses curated evidences and target systems design artifacts to design an safety case GSN from the certification criteria.}
    \label{fig:ACG}
\end{figure*}

\begin{figure}[h!]
    \centering
    \includegraphics[width=\columnwidth]{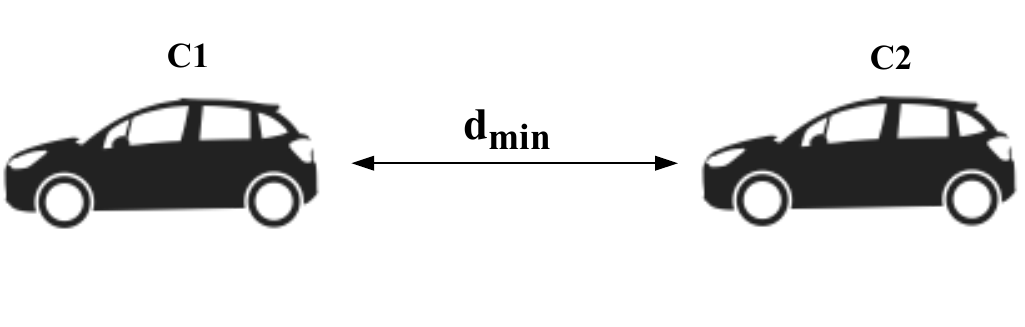}
    \caption{The AEBS case study for DeepNNCar platoon, where the follower car (C2) is required to maintain a minimum safe distance $d_{min}$ from the leader car.}
    \vspace{-0.6em}
    \label{fig:car_platoon}
\end{figure}

Despite the importance of designing robust safety case for safety-critical systems, there have been cases (explained in \cite{kelly2004systematic}) like the Clapham Rail Disaster \cite{edwards1997railway} and the Piper Alpha off-shore oil, and gas platform disaster \cite{cullen1993public} where substandard safety documentation did not confer to any standards and lacked any systematic approach by designers. These incidents illustrate the hazards of not clearly understanding, communicating and documenting a safety report. 
The main problem as seen from these incidents is the use of text (e.g, English) in the safety reports. The ambiguity of using text hinders designers or operators at different levels to generate a robust safety report. This problem of using text led to a requirement of a structured notation to explain claims, arguments and evidences. This requirement led to introduction of Claims-Argument-Evidence (CAE) and Goal Structuring Notation (GSN). 

\subsection{Goal Structuring Notation (GSN)}

Goal Structuring Notation (GSN) \cite{kelly2004systematic} was introduced by Tim Kelly at the University of York. It is a graphical argument notation which explicitly represents the elements of a safety case (claims, sub-claims, requirements, context, assumptions and evidence) as a node structure and maps the relationship that exists between each of these nodes. The purpose of this graphical goal structure is to show an iterative breakdown of the goal to sub-goals, with a mention of the assumptions and context under which the decomposition can be made. Also, the iterative decomposition continues to a point when there is no further decomposition, or we have sufficient evidence to support the parent claim. 

The GSN with its graphical representation clearly removes the ambiguities involved in defining a safety case using textual language, but this does not qualitatively imply anything about the argument itself (as explained in \cite{nair2015evidential}). The arguments, sub-goals, and evidences constituting a safety case can be imperfect, and the conventional GSN has no justification (or rationale) to indicate if the sub-goals or the evidence is sufficient to support the safety case. Uncertainty in the sub-goal's supporting evidence can lower the assurance of the entire safety case. 

To overcome the uncertainty problem in the conventional GSN, Hawkins et al. \cite{hawkins2011new} proposed a new structure of safety arguments called the assured safety arguments which extends the conventional safety argument by decomposing it into two separate arguments: (1) safety argument -- performs the decomposition of the safety goal and presents a strategy to explain the reason behind it, and (2) confidence argument -- holds the justification about the sufficiency of the confidence in the safety argument being made. This extension of the confidence arguments would provide a mechanism to backtrack the branching decisions of the GSN, while making it robust. We are currently working on extending our GSN to have a rationale block which holds a justification about the evidence used while selecting a (sub)goal.

\subsection{ALC Toolchain}
\label{sec:alc_toolchain}
The \ac{alc} Toolchain \cite{hartsell2019model} provides an integrated set of tools for development of \ac{cps} with a particular focus on systems using Learning Enabled Components (LECs). The toolchain supports \ac{gsn} safety case and allows individual nodes within an safety case to contain links to other artifacts including system architectural models, testing results and analysis, formal verification results, etc. Within a \ac{gsn} argument, these linked artifacts can be interpreted as needed based on the desired task such as justification of argument structure, supporting evidence blocks, or system functional decomposition among others. The \ac{alc} Toolchain is the future implementation platform for the automated \ac{acg} method presented in this paper and is referenced for implementation details in the remaining sections.

\section{AN ILLUSTRATIVE EXAMPLE}
\label{sec:EBS}
To illustrate our proposed ACG tool, we introduce a realistic application of an Automatic Emergency Braking System (AEBS) using an RC car testbed DeepNNCar. AEBS is a feature that will automatically override any driver input and apply maximum braking to immediately stop the car in the case of an imminent collision. To motivate the AEBS safety case, we consider the example of a car platoon as shown in \cref{fig:car_platoon}. Throughout this text, the front car will be referred to as ``Leader" and the second car is referred to as ``Follower".

These cars move around an indoor race track with varying speed ($V_t$) and steering controls ($S_t$), and are equipped with a several sensors including 2D LIDAR, IR Opto-coupler, and decawave positioning \cite{CUWB}. The entire operation of the sensing-perception-control operations of these cars are shown in \cref{fig:car_operation}. At each time step t, the sensors of the cars capture different observations $O_t$ which includes the current position $p_t$ = ($x_t$, $y_t$), current speed ($V_t$), current steering ($S_t$) and the distance to the car in front of it ($d_t$). These observations are used by different software components in the perception and state estimation block to generate information ($i_t$) which includes lanes detected ($L_t$), object detected ($Obj_t$) and slip status ($slip_t$). These values are then passed to the different controllers which calculate the required actuation commands ($a_t$) including the steering and speed Pulse Width Modulation (PWM) values.

\textbf{Certification Criteria}: To prove the AEBS reliably works, we need to assure that the follower car always maintains a minimum safe separation distance ($d_{min}$) from the leader car. This follows from the reasoning that, for $d_t \geq d_{min}$, the AEBS system has enough time to compute and apply the reverse PWM required to slow or stop the car from its current speed. Mathematically, the problem can be expressed as:

\begin{equation}
   d_{min} \geq d_t \wedge (obj) 
    \label{eqn:distance}
\end{equation}

So, if there is a car detected (obj=1) by the perception subsystem, and if the distance measurement from the LIDAR $d_t$ is always greater than the minimum safe threshold distance $d_{min}$, then we can prove in the safety case that the cars will always avoid collision under a set of assumptions. From \cref{eqn:distance} above, we see the safety case requires conjunctive claim proof that the object (car) is detected and the distance from the LIDAR $d_t$ is always greater than $d_{min}$.

Further, the first step of decomposing the certification criteria for AEBS is performed based on the state (stopped, moving) of the cars (we assume that we know the states and position of both the cars). For the two car platoon, there are four possible operating modes: \textit{mode1} - Leader stationary \& Follower moving, \textit{mode2} - Leader moving \& Follower moving, \textit{mode3} - Leader moving \& Follower stationary, and \textit{mode4} - Leader stationary \& Follower stationary. For the AEBS case study, we are particularly interested in modes 1 and 2 where the follower car is moving. A similar iterative decomposition is performed using the proposed ACG tool till the lowest level leaf node with conclusive evidence is reached. (illustrated in \cref{sec:example})

\section{The ACG Tool}
\label{sec:ACG}
\begin{figure}[t]
    \centering
    \includegraphics[width=\columnwidth]{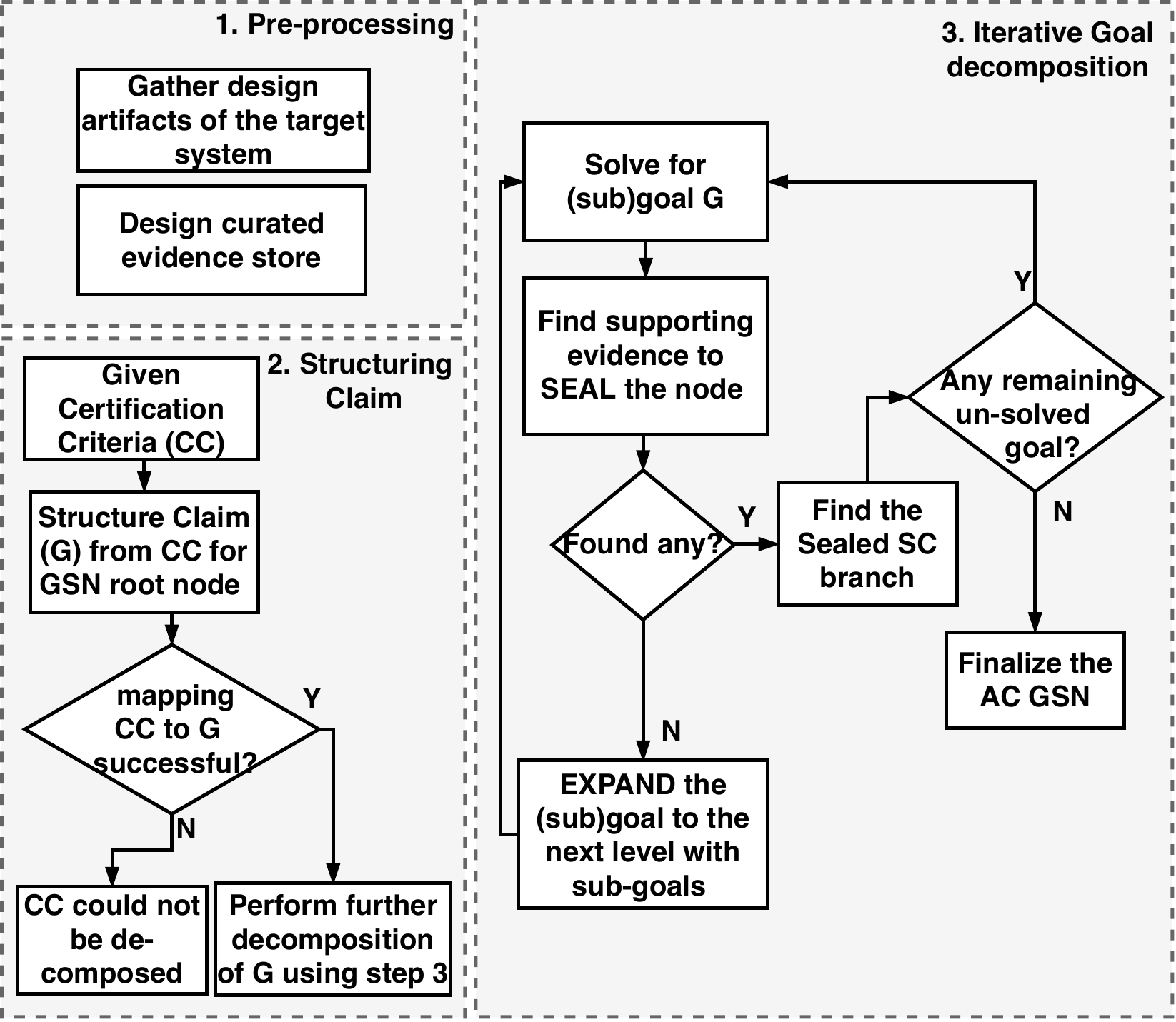}
    \caption{Phases involved in automated safety case generation using the ACG tool. 
    }
    \label{fig:flowchart}
    \vspace{-0.8mm}
\end{figure}

The Assurance case generation tool (\cref{fig:ACG}) is responsible for automatically generating the safety case for a given certification criteria, and evaluating that safety case with a confidence score (discussed in Section \ref{sec:quality}). The generated safety case will be expressed as a GSN. \cref{fig:flowchart} illustrates the three phases involved in the \ac{acg} process, they are:  


\begin{enumerate}[noitemsep,leftmargin=*]
    \item \textit{Pre-processing} involves accumulating system design artifacts and curating an evidence store about the components of the target system.
    \item \textit{Structuring Claim} involves construction of a structured claim from the given certification criteria.
    \item \textit{Iterative Goal Decomposition} involves applying decomposition operators (Link-Seal-Expand) to decompose the GSN until the leaf nodes are reached.
\end{enumerate}
The steps involved in each phase is elaborated in greater detail in the subsections below.

\subsection{Design Artifacts}
The proposed ACG tool will systematically leverage the system design information to automate the decomposition of the certification criteria to smaller sub-claims that can eventually be mapped and sealed using appropriate evidences. One way of using the entire system information is by breaking them into different graphical representations. These graphs can then be used to find supporting evidences and find relational operators among sub-claims during the safety construction process. Some of the graphs that can be used are (illustrative graphs for the DeepNNCar AEBS example is listed if available):

 \begin{itemize}[noitemsep,leftmargin=*]
    \item \textit {System Functional Breakdown (SFD)} is a logical breakdown of the system functionality that is involved or responsible for a specific mission. (\cref{fig:car_operation}-b)
    \item \textit{System Physical Decomposition} is a logical breakdown of the system into sub-systems, functions and components. (\cref{fig:car_operation}-a)
    \item \textit{Interconnectivity Graph} is a graph with interconnectivity among the different hardware and software components. (\cref{fig:car_operation}-a)
    \item \textit{Behavioral Graph} captures the activities associated with a system/component.
    \item \textit{Mapping Diagrams} are graphs which indicate the mapping of functions to components, software blocks to hardware blocks, or activities to hardware or software, etc. (\cref{fig:car_operation}-a)
    \item \textit{Ontology Graph} uses a common, domain-specific set of terms and relationships to support mapping of concepts across the other graph-of-graph elements.
    \item \textit{System Architectural Model} captures all components in the system and the interconnections between them. An architectural model constructed in the \ac{alc} Toolchain for the DeepNNCar is shown as a "Design Artifact" in \cref{fig:ACG}.
 
 \end{itemize}

\subsection{Curated Evidence Store}

The Evidence Store is a table consisting of information about the various components and functions, evidence artifacts supporting their correct operation, and all assumptions required for components to work correctly. Evidence artifacts can take any form (eg. statistical analysis of test data, analytical analysis, etc.), but must provide enough proof that the corresponding component works reliably under the stated assumptions. It is up to the developer to determine when the available evidence is sufficient for use in the safety case.

Curated evidence for the different components of DeepNNCar's operation is shown in Table \ref{Table:SafetyTable}. The supporting evidences for the components were gathered based on a number of randomized hardware and software tests (shown in \cref{fig:car_operation}) under different operating environments to evaluate when the components/tests succeed and fail. For e.g. (1) the camera module was tested under different lighting conditions and was found to work best in environments with evenly distributed lighting conditions with (800-1000) lumens of light, and (2) the LIDAR module was tested under different indoor and outdoor operating environment with different obstacles and was found to reliably and accurately work in smaller indoor rooms (due its operating range of 12 m). Similar information about different components of DeepNNCar was accumulated as evidences. 

Once we have accumulated evidence for all the systems components and compiled the design artifacts required to understand the alternatives and dependencies in the system, we can start the safety case generation process (illustrated in \cref{fig:flowchart}) that includes (1) structuring claims for the given certification criteria, and (2) iterative goal decomposition.

\begin{table*}[h!]
    \centering
    \setlength{\belowcaptionskip}{-6pt}
    \renewcommand{\arraystretch}{1.2}
       \footnotesize
    \begin{tabular}{|p{5cm}|p{1.1cm}|p{1.4cm}|p{1.6cm}|p{6.2cm}|}
    \hline
    Claim & Evidence Type & Function & Components/ \newline Subsystems & Assumptions\\
    \hline
    \rowcolor{Gray} Camera Module captures image of Leader car in range (0.1, 1)m. (G1) & H/W \& \newline S/W \newline Testing & Sensing & Camera  & 1) Light intensity above 100 lumens \newline 2) Leader car is within range (0.1,1)m \newline 3) Is powered.\\
    LIDAR Module provides distance of the obstacles in range (0,12)m and (0\textdegree,359\textdegree). (G2) & H/W \& \newline S/W \newline Testing & Sensing & LIDAR & 1) Leader car is within scan range. \newline 2) LIDAR scan motor is working. \newline 3) Is powered.\\
    \rowcolor{Gray} IR Opto-coupler Module provides RPM information in range (16,160). (G3) & H/W \& \newline S/W \newline Testing & Sensing & IR-Optocoupler & 1) It is mounted on the chassis. \newline 2) It can be occluded by plastic piece on wheel.\newline 3) Is powered.\\ 
    Lane Detection Module detects lanes and orientation of track segment. (G4) & S/W \newline Testing & Detection & Lane  \newline Detection & 1) Received frame from camera. \newline 2) LD algorithm parameters are correct. \newline 3) No light glare on track.\\
    \rowcolor{Gray} Obstacle Detection Module detects images of car in range (0.1, 1)m. (G5) & SW \newline Testing & Detection & Object \newline Detection & 1) Received frame the camera. \newline 2) Mobilenet V2 model weights are correct.\\
    Current Speed Module information is in range (0, 1)m/s. (G6) & S/W \newline Testing & State \newline Estimation & current  \newline Speed & RPM-Speed conversion is correct.\\
    \rowcolor{Gray}Current Position Module updates cars position on track. (G7) & S/W \newline Testing & State \newline Estimation & current  \newline Position & Working on indoor track with lanes\\
    Obstacle Distance Module provides distance of the obstacles in range (0,12)m (150\textdegree,180\textdegree). (G8) & S/W \newline Testing & State \newline Estimation & Obstacle  \newline Distance & Obstacle is within the scanning range (0, 12)m.\\
    \rowcolor{Gray} Slip Status Module identifies wheel slip. (G9) & S/W \newline Testing & State \newline Estimation & Slip Status & 1) Track surface is known. \newline 2) Opto-coupler module is working correctly.\\
    Steer LEC Module provides steer in range (-30\textdegree, 30\textdegree). (G10) & S/W \newline Testing & Driving & LEC Steer & 1) Receives frames from camera. \newline 2) Trained deep-learning model weights are correct. \newline 3) Trained model has seen the track before.\\
    \rowcolor{Gray} OpenCV Steer Module steer in range (-30\textdegree, 30\textdegree). (G11) & S/W \newline Testing & Driving & OpenCV  \newline Steer & 1) Receives frame from camera. \newline 2) No light glares on track. \newline 3) Lane-steer conversion is correct.\\
     Braking Manager Module provides reverse polarity $R_{PWM}$ to brake the car.(G12) & S/W \newline Testing & Braking & Braking  \newline Manager & Track surface allows braking.\\ 
    \rowcolor{Gray} Driving Manager Module provides Steer PWM in range (10,20) \& speed PWM in range (15.58, 15.62). (G13) & S/W \newline Testing & Driving & Driving  \newline Manager & Receives updated sensor and processed data (not stale ones).\\
    PWM applicator Module applies PWM at 100Hz. (G14) & H/W \& \newline S/W  \newline Testing & Driving & PWM  \newline Applicator & 1) Wiring to motors and GPIO done correctly. \newline 2) Raspberry Pi is powered.\\
    \hline 
    \end{tabular}
    \caption{The sub-goals of the different functions is shown along with its requirement, evidence type and the assumptions under which they hold.}
    \label{Table:SafetyTable}
    \vspace{-0.4em}
\end{table*}


\subsection{Structuring Claims for GSN Root Node}

The first step of generating the safety case is to design a structured claim for the root node of the GSN from the certification criteria. After this step we expect the certification criteria given in natural language to be expressed as a structured claim statement referring to elements in the design artifacts. In the case study of AEBS, the structured claim for the certification criteria (maintain safe distance and avoid collision) is expressed in \cref{eqn:distance}. This mapping of the certification criteria to the claim was manually performed by a human expert based on the states of the cars (explained in Section \ref{sec:EBS}), where the AEBS is important. Once we have a claim for the root node of the GSN, we can start an iterative decomposition process using some basic operators (Link-Seal-Expand), and logical connectives.

\subsection{Core Steps for Decomposing GSN Goal}
After obtaining a structured claim for the root node of the GSN, it must be further decomposed into sub-goals (and sub-claims) until a leaf node is reached. In this work, we define leaf nodes to be component-level nodes which have direct supporting evidence found in the curated evidence store. The decomposition in the ACG tool is handled by three primary operations Link, Seal and Expand, along with basic logical connectives. 

\subsubsection{\textbf{Link}}
\label{sec:link}
To avoid reuse of evidence for similar GSN branches, we use the link operation to link evidence nodes that have been previously used in different branches of the GSN. Effectively, if the link has seen a relationship proved previously among the different sub-goals, then this information can be used in the future to stop the exploration for evidence. This feature of the link is helpful in reducing the required number of iterations (explained for the DeepNNCar example in Section \ref{sec:example}). 

For this, we use Link as a background step which involves creating an evidence repository with modular evidences for various claims which can be reused in different GSN branches or safety cases. As discussed above, the construction and management of the curated evidence store itself can be partially automated. However, manual input from the developer is still required to link each source of evidence to the relevant system components and safety claims. 

\subsubsection{\textbf{Seal}}
The seal operation queries the curated evidence store to look for supporting evidences of a (sub)goal, and then decides if the available evidence is sufficient to stop the iteration or further evidence is required. Evidence is said to be sufficient if the claim in the (sub)goal cannot be further decomposed and there is evidence available to directly support the claim. Some sub(goals) may have evidence to support the claim, and in such cases the iteration looking for supporting evidences can be stopped. However, in cases involving higher level goals, the available evidence may not be sufficient to directly support a claim. In this case, it is necessary to query the evidence store repeatedly until sufficient evidence is available. For goals with supporting evidence from the store, an evidence with maximum confidence (explained in Section \ref{sec:quality}) that satisfies the goal under the mutual satisfaction of the assumptions is selected. Sometimes it is also possible that a sub-goal may not have sufficient evidence, in such cases we claim the node to be \textit{orphaned}. Whenever an orphaned node is found, we can stop the search and prove the safety claim cannot be argued until new evidence to support the node is available.

If the Seal operation has found a linking evidence (from the evidence store or Link operation), then it seals of the sub-goal without allowing for further exploration. A component may have multiple supporting evidences based on the operating conditions and its functionality. In such cases the seal node will have to select one evidence artifact from the available options, and this selection process is discussed in \cref{sec:quality}.

\subsubsection{\textbf{Expand}}
When no evidence for directly supporting a goal is found during the \textit{Seal} phase, the goal must be decomposed into multiple sub-goals with corresponding assumption, context, and strategy nodes. The expand operation drives this decomposition of goals into sub-goals using the available design artifacts. Each goal in the generated safety case corresponds to a system function which may require inputs from other components in order to operate correctly. The system architectural model in \cref{fig:car_operation} shows these dependencies between components, and is one design artifact used to drive goal decomposition. The strategy which connects the sub-goals can be formalized using different logical combination functions. This step refines the argument strategy of the decomposed goal node to find an appropriate logic gating function.

\textbf{Logical Connectives (gating functions)}: The decomposition of the GSN results in the goal node being split into sub-goals that can be connected using different gating functions including AND and OR. For the AEBS example, the claim resolution step results in two scenarios Mode 1 \& 2, which can be connected by an OR operator as shown in \cref{fig:safety-graph}. This decomposition is performed based on the states of the cars (explained in Section \ref{sec:EBS}). A similar illustration of using the gating function to combine the sub-goals is shown in \cref{fig:safety-graph} (explained in \cref{sec:example}).

\subsection{Implementation and Automation}
\label{sec:implementation}

The \ac{alc} Toolchain introduced in Section \ref{sec:alc_toolchain} will be used to implement the \ac{acg} including tool automation where appropriate. The following paragraphs explain how each step in the \ac{acg} workflow, shown in \cref{fig:flowchart}, will be automated.

The \textit{Pre-Processing} step consists of accumulating the available design artifacts and constructing an evidence store. For systems designed using the \ac{alc} Toolchain, all design artifacts are automatically cataloged in a version-controlled database and may be cross-referenced from other models as needed, effectively eliminating the need for a developer to collect artifacts manually. Systems designed outside of the toolchain may upload design artifacts to be added to the catalog. Construction of the curated evidence store itself can be partially automated with use of this catalog. All artifacts in the catalog which represent sources of evidence for the safety case (eg., results from system testing, formal verification, or any user-defined analysis methods) can be used to automatically populate the available evidence in the evidence store. However, the developer must manually link each piece of evidence to the corresponding component in the system architectural model, determine when the available evidence is sufficient to support a claim, and define any assumptions required for the evidence to be valid.

Next, the root node of the safety case must be derived from the certification criteria during the \textit{Structuring Claim} step. This step requires translating a claim about the system specified in natural language into a formal assurance claim. While there is a significant body of research on automation for such tasks, this is outside the scope of the \ac{acg} tool presented in this paper. Instead, this step is not automated and must be completed manually by the developer.

\begin{figure*}[t!]
    \centering
    \includegraphics[width=0.9\textwidth]{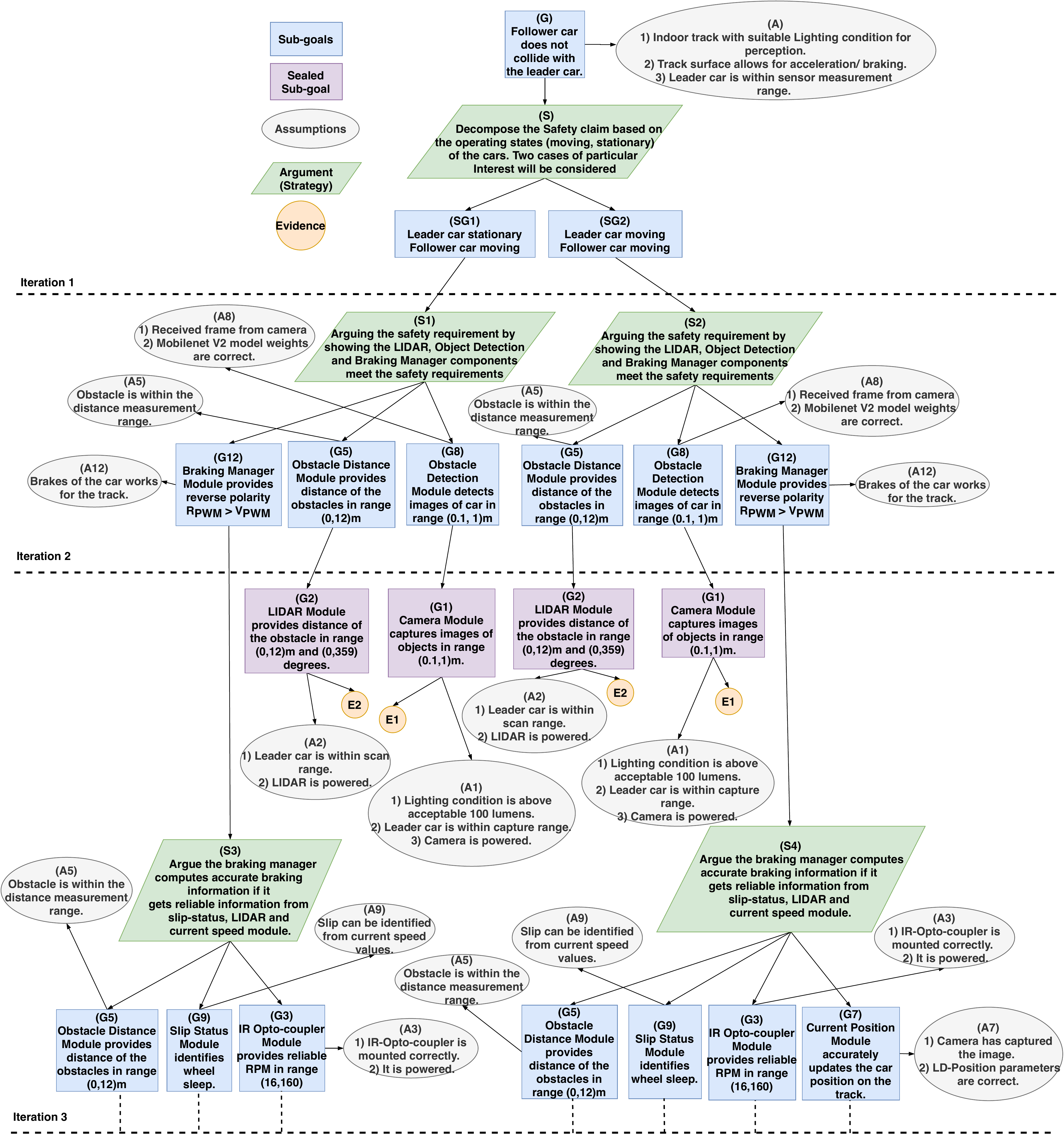}
    \caption{A GSN fragment of the DeepNNCar's AEBS case study.}
    \label{fig:GSN}
\end{figure*}

The \textit{Iterative Goal Decomposition} step interprets the available artifacts to drive the \textit{Link}, \textit{Seal}, and \textit{Expand} operations. The \textit{Link} step can be fully automated as a background task which is executed after each iteration of the \textit{Seal} and \textit{Expand} operations. When a branch of the safety case has been fully decomposed into sealed leaf nodes, this completed section can be added to the evidence store as a \ac{gsn} fragment. This way, if the fragment appears again in another branch of the safety case, it can be immediately retrieved from the evidence store instead of repeating the iterative decomposition process. The \textit{Seal} operation may be automated as a straight-forward query of both the evidence store and any \ac{gsn} fragments found by the \textit{link} operation. The \textit{Expand} operation may also be automated with an appropriate model interpreter. Since each goal is linked to a component in the system architecture model, a graph-traversal algorithm can determine all components required for a goal to function correctly. This information, combined with the assumptions listed in the evidence store, is sufficient to decompose each goal into progressively smaller sub-goals. Finally, the iterative decomposition process using the \textit{Seal} and \textit{Expand} operations can be automated with an appropriate workflow.

Once a complete safety case has been constructed, automated evaluation and correctness checks can be implemented with an appropriate formal specification language (eg. FORMULA \cite{jackson2013formula}).

\section{ACG FOR THE AEBS CASE STUDY}
\label{sec:example}
In this section we manually design a GSN for the AEBS example to outline the complexity of graphing and to illustrate the manual dependence in the design process. We then apply the ACG tool for the same example and iterate through the building blocks to generate an automated GSN. 

\subsection{Manual GSN tree generation}

A GSN fragment for the DeepNNCar's AEBS example is shown in \cref{fig:GSN}. We manually designed the GSN using the evidence store (\cref{Table:SafetyTable}) and the design artifacts (\cref{fig:car_operation}-a and \cref{fig:car_operation}-b). To prove AEBS reliably works, we need to assure the claim has sufficient evidences in both Mode 1 \& 2. So, we perform a parallel decomposition of the two modes in \cref{fig:GSN}. The conventional GSN uses different shapes for representing the blocks. In this hierarchical decomposition, the top goal is referred to as the parent node, and the decomposed sub-goals are referred to as child nodes. The principal symbols used in the GSN construction are:
\begin{itemize}[noitemsep,leftmargin=*]
    \item Blue blocks represent the goals/sub-goals, e.g. follower car does not collide with the leader car. Each blue goal is decomposed until there is direct supporting evidence from the evidence store.
    \item Gray blocks represent the various assumptions made under which the goals are satisfied. For the AEBS example, the top goal is satisfied under assumptions including that the track surface allows for permissible operation of the car, the lighting is sufficient for the sensors to work, etc. The assumptions of the parent goal is a superset of the assumptions made by the child nodes.
    \item Green blocks represent the strategy that will be used to prove that the goal holds.
    \item Orange blocks represent the evidences used to seal a goal. These evidences can be results of various randomized hardware or software tests.
    \item Purple blocks represent the sealed goals. i.e. goals which have been directly supported by evidence and further exploration is not possible or necessary.
\end{itemize}
In \cref{fig:GSN}, the different decomposition levels are separated using dotted lines, and they are termed as iterations. We can also see a few sub-goals are sealed with evidences (represented by purple blocks), while some still require further exploration (vertical dotted line extensions from the sub-goals in Iteration 3). 

Designing this GSN for a complex system like DeepNNCar is time consuming and required us to iterate through the different goals and sub-goals. To avoid the hassle of manually graphing the large GSN structure that was spread across multiple pages, we applied the ACG tool introduced in Section \ref{sec:ACG} to generate the GSN for the same AEBS example.

\begin{figure}[t!]
    \centering
    \includegraphics[width=\columnwidth]{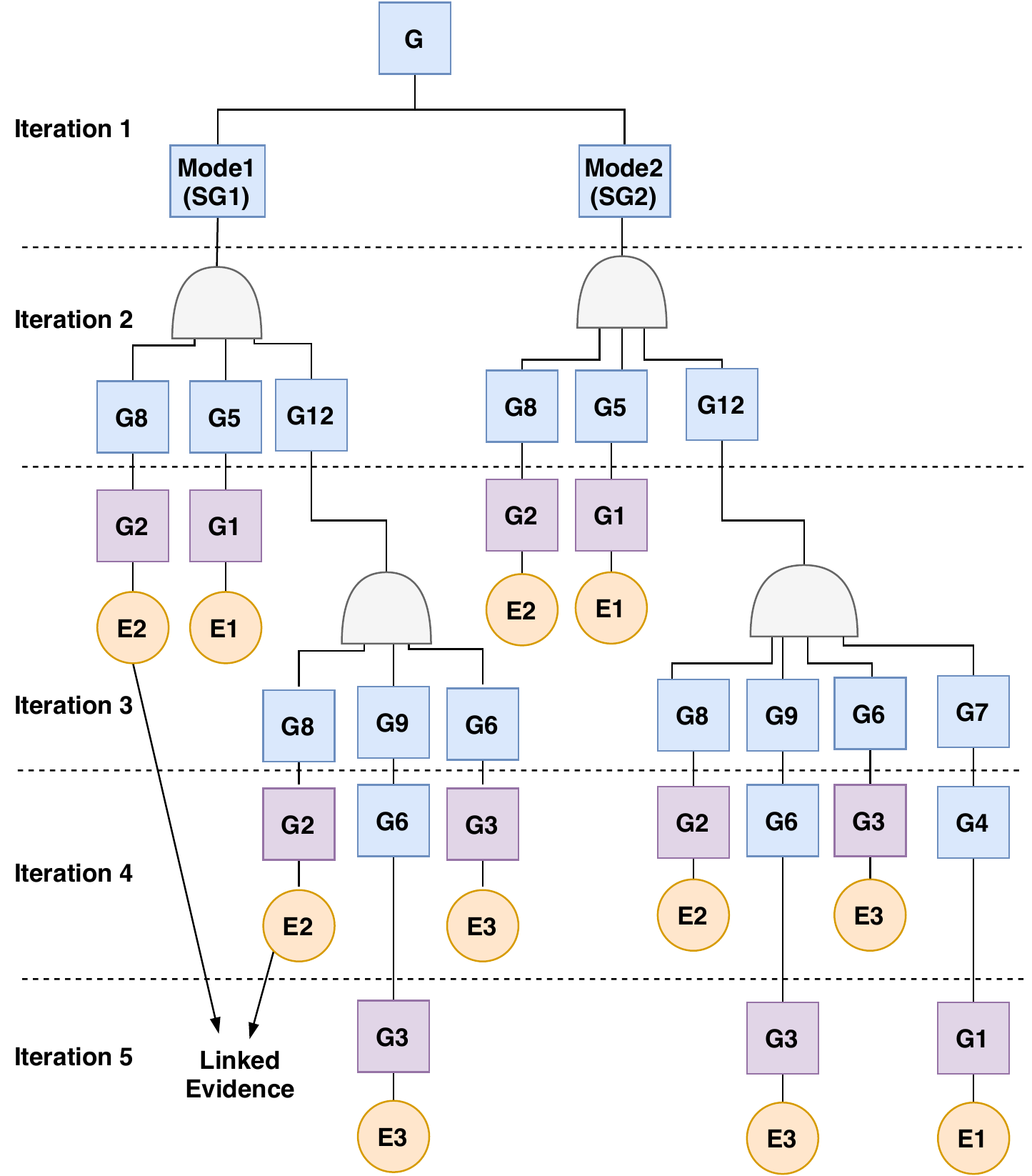}
    \caption{The graph which is generated by continuously iterating through the Link-Seal-Expand steps of the ACG tool.}
    \label{fig:safety-graph}
    \vspace{-0.6em}
\end{figure}

\begin{figure*}[t!]
    \centering
    \includegraphics[width=0.8\textwidth]{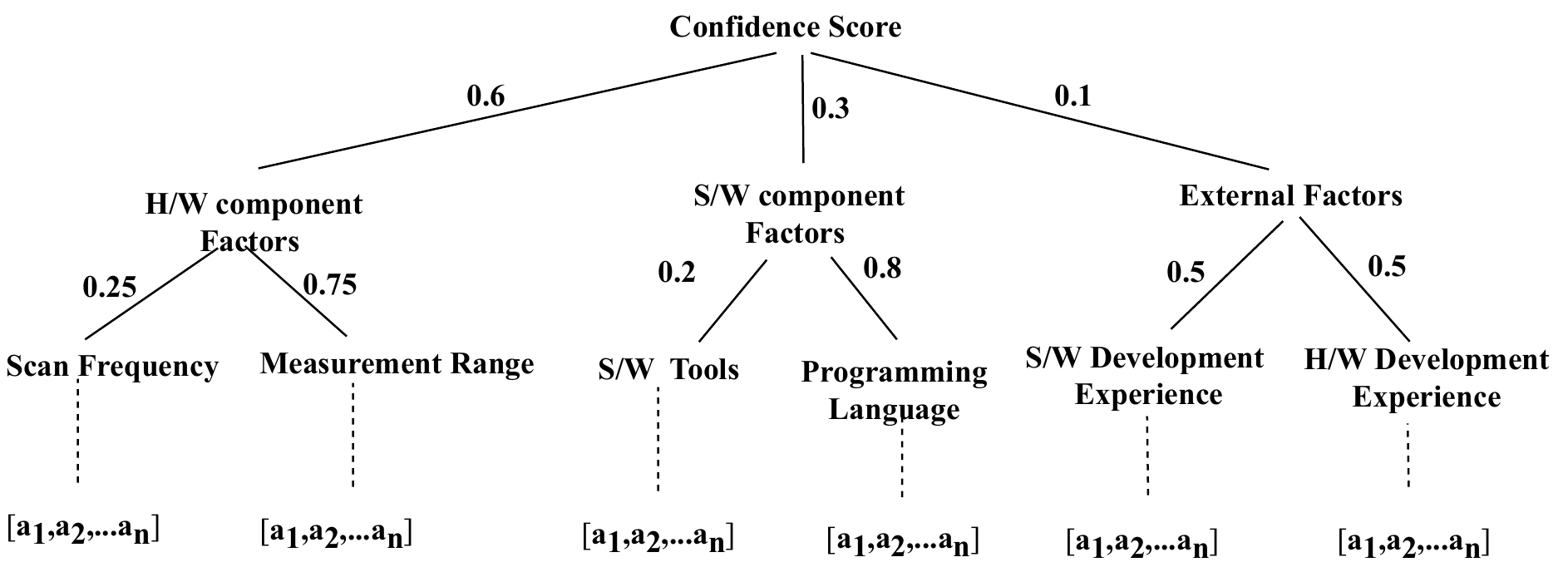}
    \caption{The evidence evaluation tree for the LIDAR module of DeepNNCar.}
    \label{fig:confidence-lidar}
\end{figure*}

\subsection{GSN generation using ACG tool}
\label{sec:automated}
The safety case generation for the DeepNNCar's AEBS example using ACG tool is shown in \cref{fig:safety-graph}. For each of the modes identified earlier we generate a safety case iterating through Link-Seal-Expand steps discussed in Section \ref{sec:ACG}. The design artifacts (\cref{fig:car_operation}-a and \cref{fig:car_operation}-b) are used along with the evidence store (Table \ref{Table:SafetyTable}) to generate the GSN.

\textbf{Iteration 1}: As the first iteration for the GSN generation, the evidence store is curated by the ALC Toolchain using design artifacts and tests on the components of the system. Once, the evidence store is designed, the claim for the root node of GSN is structured for the AEBS certification criteria by a human expert. Then based on the states of the car, the claim is parallelly decomposed into Mode 1 \& 2 as shown in \cref{fig:safety-graph}. 
    
\textbf{Iteration 2}: \textit{Seal}-- ACG tool looks for supporting evidence for the goal (AEBS) from the evidence store. Since no supporting evidence can be found in the evidence store, the \textit{Expand} operation is performed to decompose the goal. The AEBS node makes use of the functional complimentary pattern which shows that maintaining safe distance, including stopping when obstacle distance is less than $d_{min}$, requires the conjunction of three functions: obstacle detection, measuring obstacle distance, and the breaking manager providing appropriate PWM signal. This results in the three sub-goals \{G8, G5, G12\} logically connected using the AND operator with the parent AEBS goal. Since there is no concluding evidence from the current leaf nodes \{G8, G5, G12\}, the ACG tool iterates this step again to decompose these sub-goals further.
    
\textbf{Iteration 3}: \textit{Seal}-- ACG tool looks for supporting evidences for the sub-goals \{G8, G5, G12\} from the evidence store, since no supporting evidence can be found in the evidence store, the \textit{Expand} operation is performed to decompose the goals. Looking at the evidence table, and functional breakdown graph (\cref{fig:car_operation}-b) the ACG tool performs the following goal decomposition's: (1) obstacle distance node (G8) depends on the LIDAR node (G2), (2) the obstacle detection node (G5) depends on the camera node (G1), and (3) the braking manager node (G12) depends on the slip status (G9), IR-Optocoupler (G3), Object distance (G8) and current position (G7) nodes (Only for the case when the two cars are moving). The sub-goals of the braking manager are logically connected using the AND operator. Again, since no evidence was found for the three sub-goals, ACG tool iterates further.
    
\textbf{Iteration 4}: \textit{Seal}-- Again, the ACG tool looks for evidences of sub-goals \{G2,G1,G8,G9,G6,G7\} from the evidence store to seal off the branches. The tool finds, conclusive evidences for G1 and G2 and seals them. In this work we advocate the selection of an evidence with maximum confidence score. However, most of the component sub-goals of this example has only one piece of supporting evidence for selection. So, we directly select the only available evidence to seal the nodes. All the other sub-goals of G6, G7, G8 and G9 do not have supporting evidence and must be further expanded. \textit{Expand}-- the obstacle distance node (G8) depends on the evidence from LIDAR node (G2), the slip status node (G9) depends on evidence from current speed node (G6), and (G6) depends on evidence from the IR-Optocoupler node (G3), and the current position node (G7) depends on evidence from Lane detection node (G4) (only for Mode2, as in this case both the cars are moving and the braking manager requires constant information updates of the position of the cars, which is not that important in Mode1 as the leader car is stationary). Since, none of these are nodes are sealed, a further iteration is performed by the tool. 
    
\textbf{Iteration 5}: \textit{Seal}-- The ACG tool again looks for evidences of sub-goals \{G6, G3, G4(for mode 2)\} from the evidence store and finds that node G3 has supporting evidence. Also, the current speed node (G6) has dependence on G3, for which evidence has already been found the evidence store. So, using the evidence of G3 as linking evidence, the tool seals both sub-goal nodes. However, G4 has no supporting evidence and hence is further expanded. \textit{Expand}: Lane detection node (G4) is further decomposed to a single node G1, and a further iteration is performed to find an evidence for it.
    
\textbf{Iteration 6}: \textit{Seal}-- the ACG tool looks for the evidence of the camera node (G1) to seal it off. Since it is a component node, direct evidence is available from the evidence store. Since every branch of the GSN is sealed off with evidence, iteration is finished, and the AC is complete.

\section{Safety CASE EVALUATION}
\label{sec:quality}

Safety case evaluation is important for quantifying the confidence of the generated safety case \cite{denney2011towards}. For evaluating the credibility of the generated safety case, we compute a confidence score to the top goal node associated with the certification criteria. Our mechanism to identify this score is based on a bottom up approach, which moves the confidence associated with the evidence used to seal a claim. The safety evaluation in the ACG tool is now performed manually by a human expert at design time, but we are working on automating it.

\begin{table}[t]
    \centering
    \setlength{\belowcaptionskip}{-6pt}
    \renewcommand{\arraystretch}{1.2}
       \footnotesize
    \begin{tabular}{|p{2.5cm}|p{3.5cm}|p{1cm}|}
    \hline
    Attribute & Assessment & Score\\
    \hline
    \rowcolor{Gray}Scan Frequency & precise at 2Hz \newline Precise at 4 Hz & 0.4 \newline 0.6\\
    Measurement Range & precise in range 0-3 m \newline precise in range 3-6 m \newline precision in range 6-12 m  & 0.8 \newline 0.6 \newline 0.5\\
    \rowcolor{Gray}Software Tools & -- & --\\
    Programming Language & Python  \newline C++ & 0.5 \newline 0.8\\
    \rowcolor{Gray}Software Development Experience & Programming Knowledge \newline  Communication protocols Knowledge & 0.4 \newline 0.7 \\
    Hardware Development Experience & PWM Knowledge \newline Motor Knowledge \newline UART Knowledge & 0.8 \newline 0.6 \newline 0.7\\
    \hline 
    \end{tabular}
    \caption{The assessment score Table for attributes of the LIDAR component of DeepNNCar.}
    \label{Table:Assessment-Table}
    \vspace{-0.4em}
\end{table}

\subsection{\textbf{Confidence Score Estimation}}

Based on the operating context and the expert's assessments regarding the component module, every evidence node in the GSN structure gets a confidence score, and the ones with the higher score will be chosen as the supporting evidence. The confidence score can be computed using an approach described in \cite{nair2015evidential} which performs confidence evaluation using Evidential Reasoning (ER) \cite{yang2002evidential}. ER is a concept of assimilating multiple attributes of a piece of evidence into a single coherent assessment. Specifically, the evidences of a claim are decomposed into various different attributes $\{e_1,e_2,......e_n\}$ which are then further decomposed to sub-attributes and this process is repeated till further breakdown is not possible. We refer to this structure as the evidence evaluation tree, and we have designed one such structure for the LIDAR module of DeepNNCar (see \cref{fig:confidence-lidar}). The attributes and sub-attributes of the evidence tree vary according to the class of evidence (software module, hardware module) and the context of operation. 

At the lowest level of the attribute decomposition, the designer can provide a score for the different assessment $\{a_1,a_2,......a_n\}$ regarding the attribute. These assessments vary for different class of evidences (software, hardware) and the context of their operation. A sample attribute assessment for the LIDAR module is shown in Table \ref{Table:Assessment-Table}. Every attribute can have an assessment and a score, which is statistically computed by a human expert based on his experience of how the components work under different scenarios. In Table \ref{Table:Assessment-Table} we have a few assessment scores for different attributes of a LIDAR module, and from various tests of the component we assigned the score. We found the LIDAR to most precisely measure distances of objects in range 0-3 meters, so a score of 0.8 is assigned, and similarly the LIDAR's precision degrades when the object is in the range 6-12 meters, and so a score of 0.5 is assigned. A similar assessment score was evaluated for each of the other attributes. These assessment scores are also referred to as belief functions.

Once, we have the belief function for all the leaf level sub-attributes, and if the importance (weights) of the attributes is available towards the claim, then the ER algorithm \cite{yang2002evidential} can be used to assimilate them. The algorithm is developed based on multi-attribute evaluation framework and evidence combination rule of Dempster–-Shafer (D–S) theory of evidence \cite{krause_1991}. The three steps involved in the algorithm (as explained in \cite{jiao2019analysis}) are: (1) weighting the belief distribution -- weights are assigned to the belief distribution based on the importance of attribute towards the top goal safety claim, (2) aggregation process -- combine all the assessment of the basic/ sub-attributes, and (3) generation of combined belief degree -- after aggregating the assessments for all the basic attributes, the combined belief degree is computed for the entire evidence.

The weights of the attributes are important in computing the confidence score, and they can be estimated either randomly by the designer based on each attribute's importance towards the goal, or can be computed using elaborate methods of pairwise comparison of attributes \cite{tzeng2011multiple} (In this work the weights are chosen by the expert based on his intuition of the component). Then the beliefs are propagated from the leaf nodes, combined along with the weights, and summed with the scores of the other sibling sub-attributes to compute an assessment score. This assessment score represents the overall confidence of the evidence. A similar evaluation of the evidence can be performed for all the nodes in the GSN structure. 

Once a confidence score is available on all the sealed evidence nodes, we use compound semantics based upon the logical operators used in the strategy combining the sub-claims. Some similar composition using logical operators has been applied in literature \cite{pradhan2016achieving, nannapaneni2016mission} for reliability estimation in CPS. The composition works as follows:
\begin{itemize}[noitemsep,leftmargin=*]
    \item AND operator will propagate the minimum confidence score.
    \item OR operator will propagate the maximum confidence score from all the available branches. 
\end{itemize}

The safety case evaluation scheme will assign a confidence score to the top goal node associated with the certification criteria.

\subsection{\textbf{Evidence Coverage}}

We are also currently working on integrating the metric of evidence coverage as one of the attributes used in evidential reasoning. From the GSN structure we can infer that the claims supported by the evidence should always be a superset of the claims made in the goal node. Also, the assumptions made by the evidence should be a subset of the assumptions made in the context of the goal node which we are trying to seal with the evidence. We use this containment relationship among the GSN blocks at different hierarchical levels to evaluate the quality of the supporting evidence. We term this method of evaluating the evidence based on their containment relation to the higher-level goal as ``Evidence Coverage". We use this based on our hypothesis that higher score should be given to the evidence that provides the biggest margin between the assumptions. Also, this metric is qualitative unlike the confidence score which is quantitative.

\section{Related Work}
\label{sec:related_research}

As explained in \cite{rinehart2015current}, current safety case construction practices can be divided into one of three categories: \textit{Prescriptive} where standards explicitly define the required development processes and procedures, \textit{Goal-Oriented} where high-level safety goals are specified but the process for achieving them is flexible and left to the system developer, or \textit{Blended} which uses aspects from both of the other categories. Rinehart et al. examine the processes used in various industries and show that prescriptive techniques tend to be used in industries with well understood technologies and a history of safe operation. However, they note there is a general trend toward goal-oriented approaches, similar to the ACG tool proposed in this paper, such as the Risk-Informed Safety Case \cite{dezfuli2011nasa} from NASA. Goal-oriented approaches appear to be a suitable option for CPSs which operate in highly uncertain environments.

There are several commonly encountered pitfalls in the construction of safety cases. Leveson \cite{leveson2011use} identifies a variety of these pitfalls and provides suggestions for avoiding them. For one, safety case construction and system safety analysis should be an ongoing process started early in the design cycle as opposed to a discrete activity performed near the end of system development for the purpose of certification. Leveson also shows that safety cases are prone to confirmation bias and argues that developers should instead attempt to show when a system can become \textit{unsafe}. "The Nimrod Review" \cite{haddon-cave2009nimrod} provides examples of both of these fallacies where the Nimrod aircraft was inherently assumed to be safe due to a history of safe operation, and the resulting safety case did not provide any real improvement in the safe operation of the aircraft. While some of the lessons learned from these examples require human input and understanding to address, other issues can be mitigated with the use of appropriate assurance case tools. These include enforcing the use of sound safety case patterns, promoting early and continuous development by reducing the time required for construction, and tightly integrating system assurance with the relevant system models and documentation among others.

For simplifying the ACG process several commercial and research tools are developed. Maksimov, Mike, et al. \cite{maksimov2018two} provide a comprehensive survey on the assurance and safety case tools developed in the last two decades. This paper reports 46 assurance case tools and evaluates them based on their capability to generate, maintain, assess and report safety cases. For comparison we have listed a few commercial and research tools. Commercial tools mainly focus on providing a platform for developing and managing assurance cases. Assurance Case Construction and Evaluation Support System (ACCESS) \cite{steele2011access} is a tool based on Microsoft visio that aids in creation and maintenance of safety cases. It provides a platform for rapid prototyping, node creation, node coloring, of the GSN argument structure it creates. CertWare workbench \cite{barry2011certware} is another tool based on Eclipse that provides various functionalities like multi-user safety case editing, change tracking, standard safety case templates, and cheat sheets for simple and fast safety case development. Similarly, Assurance and safety case environment (ASCE) \cite{adelard2011assurance} provides an environment for simple safety case creation and management, and allows for simple and low cost generation of safety case reports. Also, D-Case editor from DEOS \cite{matsuno2011d} is an open-source platform implemented as an eclipse plugin to generate and manage GSN argument structures.   

In addition to these, several research tools have made significant improvements in automating the safety case generation process. Gacek et al. \cite{gacek2014resolute} introduce Resolute, a tool which generates safety cases from system architecture models specified in AADL \cite{feiler2006architecture} along with formal claims and rules specified in an appropriate domain-specific modeling language. Resolute can also automatically propagate updates from the architectural model to the safety case and check for any assumption violations, but manual effort is still required to construct the formal claims and rules. Similarly, Calinescu et al. \cite{calinescu2017engineering} apply ACG techniques to self-adaptive systems with the ENTRUST methodology. This approach generates dynamic assurance cases which adapt along with the system to remain valid after system reconfiguration. Additionally, Denney et al. \cite{denney2018tool}, examine several such tools and provide an introduction to their AdvoCATE toolset. AdvoCATE introduces a methodology for automated generation of safety cases and provides functionality for argument analysis and improvement, evidence selection, and claim definition and composition.

\section{DISCUSSION}
\label{sec:discuss}
In this section, we first discuss the functionalities of the ACG tool, which was motivated because of the limitations in the existing ACG process and tools. We then discuss the future enhancements to the automation of our ACG tool.

\subsection{Reflections}
As discussed before, large efforts are being made in designing safety case reports for safety-critical systems which has recently become a mandate in many industries. Despite significant efforts and improvements, developing these reports are typically a manual process requiring human involvement at several steps. This has made the ACG process slow and error prone. Overall, we feel less attention is being paid to the methods by which these safety case reports are being developed. This was the primary motivation behind the proposal of the ACG tool which significantly reduces the cost of assuring \acp{cps}, provides a robust process which minimizes human involvement, and accelerates the entire process of safety case generation. Specifically, the functionalities of the ACG tool that were motivated by the limitations in the existing ACG process and tools are:

\begin{itemize}[noitemsep,leftmargin=*]
\item Automated artifact management (\cref{sec:implementation}) and partially automated evidence store generation from the system architecture models using the ALC Toolchain (\cref{sec:alc_toolchain}).
\item Automated safety case GSN construction (\cref{sec:automated}) using domain artifacts and the curated evidence store.
\item Linked atomic evidence nodes (\cref{sec:link}) to minimize redundancy and promote reusablability for other safety cases.
\item Safety case evaluation (\cref{sec:quality}) to determine credibility of the generated safety case using a confidence score.  
\end{itemize}

Through we have not comprehensively discussed the specific research methods (e.g., FORMULA) that have been used in the proposed tool, we believe enough description of the research methods to achieve automation is discussed in \cref{sec:ACG}. To the best of our knowledge, several functionalities (e.g. safety case evaluation) provided by our ACG tool are either not provided or primitive in most of the tools discussed in \cref{sec:related_research}.

\subsection{Future Work}

As future steps, we plan to improve the existing components and enhance the automation capability of the ACG tool. Some possible extensions and improvements are listed below:

\begin{itemize}[noitemsep,leftmargin=*]
    \item Automating Claim Structuring: Currently, the conversion of the informal certification criteria into the GSN root node goal is performed by a human expert, however, as an extension we would like to use a natural language processing technique such as keyword matching \cite{ghosh2016arsenal} to automatically extract and map the informal certification requirements to goal of the GSN root node. 
    
    \item Safety Case Evaluation: Currently, confidence score is used to evaluate the generated safety case. However, confidence as a metric is probably not sufficient to evaluate the credibility of the generated safety case. To strengthen this, we may also need to evaluate the ACG in terms of soundness and stability.
    
    \item Automating Seal operation: As discussed in \cref{sec:implementation}, the steps of linking each piece of evidence to the corresponding component in the system architectural model, determining when the available evidence is sufficient to support a claim, and defining any assumptions required for the evidence to be valid are all done manually by the developer. We want to automate this process.
    
    \item Extending GSN notation: The existing automated GSN notation does not have a means to explain the justification that indicates whether the sub-goals or evidence are sufficient. This is a vital piece of argument justifying the GSN branching and supporting evidence. We are working to add a justification node to the existing GSN.
    
    \item Extending Logical connectives: Currently, the tool only supports the AND and OR logical connectives. We are working on extending to other logical connectives like XOR.  
\end{itemize}
 
\section{CONCLUSION}
\label{sec:conclusion}
Safety Case has become a part of the regulatory certification process in different CPS domains. Despite its importance, very little efforts are being made to improve the existing ACG process which is mostly been manual, ad-hoc without a systematic approach. There are tools designed to support the ACG, however, they still facilitate manual generation of safety case. To address this, we have proposed an ACG tool that considers the certification criteria along with system's design artifacts and evidences accumulated by human experts to generate a fully decomposed GSN in an automated manner. Specifically, the ACG tool along with the ALC Toolchain can automatically generate design artifacts from the system model architecture, and further populate an evidence store that is required for the safety case generation process. In addition, it can iteratively decompose the root node goal (or certification criteria) of the GSN to automatically construct a GSN using the Link-Seal-Expand steps. This automated GSN construction significantly reduces time and human effort. Additionally, the ACG tool has the capability to evaluate the generated safety case using a confidence score. This evaluation mechanism is novel and extends the existing popular tools like ASCE  \cite{adelard2011assurance} and AdvoCATE \cite{denney2018tool}. 

We also envision our tool to reduce the time and cost of the certification process, and reduce the ambiguity in a safety case that is otherwise introduced by too much involvement of human experts. Further, we have also illustrated the proposed ACG tool on an AEBS case study using a RC car testbed. Currently the ACG tool is not fully functional for online validation. We are working on integrating the different components together, so that it can be validated with other CPS testbeds. We eventually want to integrate the ACG tool into the ALC toolchain to build a single comprehensive toolchain for offline design, development and safety case generation of \ac{cps} applications.

\section*{ACKNOWLEDGEMENTS}
This work was supported in part by DARPA's Assured Autonomy project and Air Force Research Laboratory. Any opinions, findings, and conclusions or recommendations expressed in this material are those of the author(s) and do not necessarily reflect the views of DARPA or AFRL.

\balance

\bibliographystyle{IEEEtran}
\bibliography{main.bib}

\end{document}